\documentclass[10pt,twocolumn,letterpaper,pagenumbers]{article} 

\usepackage{iccv}              
\usepackage{multirow}

\usepackage{makecell}
\definecolor{iccvblue}{rgb}{0.21,0.49,0.74}
\usepackage[pagebackref,breaklinks,colorlinks,allcolors=iccvblue]{hyperref}
\hypersetup{urlcolor=magenta}
\usepackage[accsupp]{axessibility}  
\usepackage[para]{footmisc}  

\def\paperID{} 
\def\confName{ICCV}
\def\confYear{2025}

\title{HumanSAM: Classifying Human-centric Forgery Videos \\in Human Spatial, Appearance, and Motion Anomaly}

\author{Chang Liu$^{1}$\thanks{Equal contribution} \quad 
Yunfan Ye$^{2}$\footnotemark[1] \quad 
Fan Zhang$^{1}$ \quad
Qingyang Zhou$^{1}$ \quad
Yuchuan Luo$^{1}$ \quad
Zhiping Cai$^{1}$\thanks{Corresponding author} \\
$^{1}$ National University of Defense Technology\quad 
$^{2}$ Hunan University \\ \\
{\tt\small \url{https://dejian-lc.github.io/humansam/}}
}

\usepackage{xcolor}
\usepackage{graphicx}

\begin{document}
\maketitle  
\begin{abstract}
Numerous synthesized videos from generative models, especially human-centric ones that simulate realistic human actions, pose significant threats to human information security and authenticity. 
While progress has been made in binary forgery video detection, the lack of fine-grained understanding of forgery types raises concerns regarding both reliability and interpretability, which are critical for real-world applications. To address this limitation, we propose HumanSAM, a new framework that builds upon the fundamental challenges of video generation models. Specifically, HumanSAM aims to classify human-centric forgeries into three distinct types of artifacts commonly observed in generated content: spatial, appearance, and motion anomaly.
To better capture the features of geometry, semantics and spatiotemporal consistency, we propose to generate the human forgery representation by fusing two branches of video understanding and spatial depth. We also adopt a rank-based confidence enhancement strategy during the training process to learn more robust representation by introducing three prior scores.   
For training and evaluation, we construct the first public benchmark, the Human-centric Forgery Video (HFV) dataset, with all types of forgeries carefully annotated semi-automatically. 
In our experiments, HumanSAM yields promising results in comparison
with state-of-the-art methods, both in binary and multi-class forgery classification. 
\end{abstract}    
\section{Introduction}
\label{sec:intro}

Video generation models, particularly diffusion-based ones, are advancing rapidly, producing video content increasingly indistinguishable from reality~\cite{videoworldsimulators2024,wanx,kong2024hunyuanvideo,ma2025step,kling,gen3,yang2024cogvideox,minmax}.
However, such technological progress also poses unprecedented societal risks, particularly when generating human-centric forgery videos, potentially causing significant negative impacts (e.g., violation of personal privacy, spread of misinformation). 
The generated highly realistic videos that simulate complex human actions make the need more pressing than ever to distinguish between generated and real videos effectively. Recently, efforts have been made either in face forgery detection~\cite{wang2020cnn,tan2024rethinking,guo2023hierarchical,qian2020thinking} or binary classification~\cite{wang2023dire,sha2023fake,AIGVDet24,on-learning-multi-modal-forgery-representation-for-diffusion-generated-video-detection}, but both with limited reliability and interoperability.

\begin{figure}[t]
  \centering
   \includegraphics[width=1.0\linewidth]{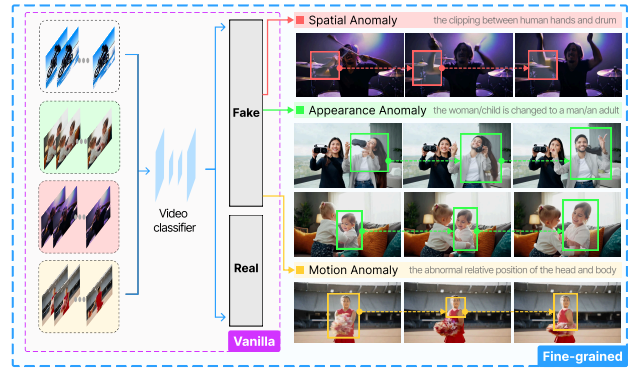}

   \caption{Vanilla video classifier distinguishes only fake or real videos, while fine-grained forgery classification further extends the fake videos into spatial, appearance and motion anomalies.
   }
   \label{fig:teaser}
\end{figure}

To address this issue, one possible solution is expanding the binary forgery video classification into multiple fine-grained classes~\cite{Xu_2024_CVPR_parser,chang2024matters,10902635}. This solution, however, relies strongly on the definition of fine-grained forgery types, which itself is a highly difficult problem to ensure the classes are exhaustive and mutually exclusive, compounded by the lack of corresponding datasets that reflect such well-defined categories. This explains why there is rarely a work on fine-grained forgery video classification even in such a video generation era. 

Recent studies identify three persistent challenges: (1) Unnatural interactions, including flawed material responses and causal inconsistencies~\cite{ma2025step,videoworldsimulators2024}; (2) Object appearance inconsistency, affecting identity preservation, object permanence, and scene coherence~\cite{si2025repvideo,yuan2024identity}; and (3) Motion fidelity issues, where models prioritize visual appearance over biomechanically plausible motion~\cite{chefer2025videojam,kong2024hunyuanvideo}. These limitations persist across architectures and scales, underscoring fundamental barriers in learning causal-temporal priors purely from visual data.

In this work, based on extensive observations and systematic analysis, also inspired by previous wisdom~\cite{chang2024matters}, we argue that human-centric forgeries can be divided into the appearance and motion of humans, and the interaction between humans and spatial objects. We define three types of human-centric anomalies—spatial anomalies, arising from unrealistic geometric interactions (e.g., hand clipping through a drum); appearance anomalies, caused by semantic inconsistencies across frames (e.g., a child transforming into an adult); and motion anomalies, reflecting unnatural or inconsistent human motion patterns (e.g., abnormal head-body alignment)—as illustrated in Fig.~\ref{fig:teaser}. To facilitate research in this issue, we introduce the first Human-Centric Forgery Video (HFV) dataset, which is designed for multi-class forgery classification.

Inspired by existing video generation benchmarks (e.g., VBench~\cite{huang2024vbench} and EvalCrafter~\cite{liu2024evalcrafter}), we accelerate the construction pipeline of the HFV benchmark by applying pre-trained vision models to help score three types of human forgeries in advance. The spatial, appearance and motion anomaly scores are mainly related to the vision tasks of depth estimation, semantics of foundation model and optical flow estimation, respectively. We then automatically assign forgery-type labels based on these scores, followed by careful manual verification to ensure the correctness of the HFV dataset. 

The calculated prior scores are also applied to the proposed rank-based confidence enhancement strategy by introducing adaptive confidence score into loss functions, to enhance the model's sensitivity to difficult samples and learn more robust representation. Based on the observation that monocular depth estimation is well-qualified to sense spatial anomaly, we propose to combine a video understanding foundation model with a monocular depth estimation model to generate a fused representation of human forgeries, enabling comprehensive and robust classification across four categories(i.e. human spatial, appearance, motion anomaly and real videos).

With the above efforts, extensive experiments on the proposed dataset show that HumanSAM outperforms other state-of-the-art methods both in binary and multi-class forgery classification, which further verified the rationality of the fine-grained forgery types. 
The main contributions include:

\begin{itemize}
\item A novel end-to-end framework for fine-grained human-centric forgery video classification, by extending the traditional binary classification to a multi-class task (i.e. three types for generated video and one for real video).

\item The first public benchmark for training and evaluating multi-class human forgeries, HFV dataset, including three forgery types of human spatial, appearance and motion anomaly, by introducing an automatic labeling and verification pipeline.

\item Several technical designs to ensure the accuracy and robustness of forgery classification, including the dual-branch fusion that integrates depth features, and the rank-based confidence enhancement strategy that integrates prior scores.

\end{itemize}




\section{Related Work}
\label{sec:related_work}

\begin{table*}[h!]
\centering
\caption{The detailed composition of the HFV dataset}
\label{tab:HAFF}
\setlength{\tabcolsep}{8pt} 
\renewcommand{\arraystretch}{0.8} 
\scalebox{0.9}{
\begin{tabular}{lccccc}
\toprule
Video Source & Quantity & Duration (s) & Frame Rate (fps) & Frame Count & Resolution \\
\midrule
MiniMax\cite{minmax} & 730  & 5    & 25 & 141 & 1280$\times$720 \\
Gen-3\cite{gen3}         & 701   & 10   & 24 & 255 & 1280$\times$768 \\
Vchitect-2.0 (VEnhancer)\cite{he2024venhancer} & 706  & 4    & 16 & 79  & 1920$\times$1080 \\
Kling\cite{kling}        & 720  & 5    & 30 & 153 & 1280$\times$720 \\
CogVideoX-5B\cite{yang2024cogvideox} & 728  & 6    & 8  & 49  & 720$\times$480 \\
Vchitect-2.0\cite{vchitect} & 725  & 5    & 8  & 40  & 768$\times$342 \\
CogVideoX-2B\cite{yang2024cogvideox}  & 720   & 6    & 8  & 49  & 720$\times$480 \\
pika\cite{pika}      & 725   & 3    & 24 & 72  & 1280$\times$720 \\
Gen-2\cite{gen2}         & 725   & 4    & 24 & 96  & 1408$\times$768 \\
K400\cite{kay2017kinetics}     & 810 & - & -&- & $\geq$224$\times$224 \\
\bottomrule
\end{tabular}
}
\end{table*}

\textbf{Video Generation Models.}  
Video generation has advanced rapidly. Early models~\cite{ho2022video,wang2023modelscope} produced short, glitchy outputs, whereas recent models like MiniMax~\cite{minmax}, Gen-3~\cite{gen3}, and Kling~\cite{kling} generate high-quality, temporally consistent videos spanning hundreds of frames. 
However, three fundamental challenges persist:  
(1) \textit{Unnatural interactions} – Models struggle with material responses and causality. Wan2.1~\cite{wanx} enhances coherence via high-order flow matching.  
(2) \textit{Object appearance inconsistency} – Maintaining identity and structural continuity remains difficult. ConsisID~\cite{yuan2024identity} injects frequency-decomposed signals into DiT, while RepVideo~\cite{si2025repvideo} stabilizes intermediate representations via feature caching and gating.  
(3) \textit{Motion fidelity issues} – Ensuring plausible motion remains challenging. HunyuanVideo~\cite{kong2024hunyuanvideo} and StepVideo~\cite{ma2025step} employ 3D full attention for stronger motion dynamics, while VideoJAM~\cite{chefer2025videojam} leverages self-generated noisy optical flow for improved motion quality.  
While existing methods focus on mitigating these flaws, we take a different approach—exploiting them to develop a human-centric multi-class forgery detector. Beyond binary classification, our method enables fine-grained anomaly detection, improving overall forgery detection performance as previous works~\cite{Peng_2024_CVPR,10665926,CHANG2021106807}.  

\vspace{0.2em}
\noindent\textbf{Forgery Detection.} 
Despite rapid advancements in video generation, forgery detection remains underexplored. Existing methods primarily target face forgeries, which fail to capture the diverse artifacts introduced by modern video generation models. These approaches can be categorized into image-level and video-level detection.  
Image-level methods detect visual artifacts in generated frames using CNN-based classifiers like CNN-Det~\cite{wang2020cnn}, DDIM inversion and reconstruction~\cite{wang2023dire} for generative artifacts, or model-specific "fingerprints"~\cite{sha2023fake}. Uni-FD~\cite{ojha2023towards} further classifies images in CLIP-ViT feature space via a nearest-neighbor algorithm. However, these methods lack temporal modeling and struggle with video-level consistency.  
Video-level detection techniques, such as frequency-domain analysis (e.g., F3Net~\cite{qian2020thinking}), two-stream networks~\cite{AIGVDet24,simonyan2014two}, and multimodal models~\cite{on-learning-multi-modal-forgery-representation-for-diffusion-generated-video-detection}, offer improved temporal analysis but remain constrained to binary classification. Moreover, most of these models require pre-extracted features, making inference computationally expensive, and have yet to be systematically evaluated on state-of-the-art video generation models.  

In contrast, we propose a human-centric multi-class forgery detection framework that categorizes generated video anomalies into three types. This finer-grained classification enhances interpretability and improves forgery detection beyond conventional binary approaches.

\vspace{0.2em}
\noindent\textbf{Video Understanding.}
Early video understanding models relied on frame-level processing, extracting information independently from each frame using architectures such as DeepVideo~\cite{karpathy2014large} and two-stream networks~\cite{simonyan2014two}. This evolved into video-level models that directly capture spatiotemporal features, including C3D~\cite{tran2015learning}, I3D~\cite{kay2017kinetics}, and TimeSformer~\cite{bertasius2021space}. With the advent of the Scaling Law~\cite{kaplan2020scaling}, large-scale video understanding models~\cite{li2023unmasked,wang2024internvideo2} and foundational visual perception frameworks such as Depth Pro~\cite{ravi2024sam,yang2024depth,bochkovskii2024depth,ye2023stedge,ye2024diffusionedge}, have achieved significant advancements.  
Appearance and motion features are essential for distinguishing normal from anomalous video patterns~\cite{Zhang_2024_CVPR}, yet depth estimation uncertainty adversely affects human pose estimation~\cite{Xu_2024_CVPR_pose}. Additionally, current video understanding models struggle with spatial logic reasoning~\cite{yang2024thinking}.  

Inspired by these insights, we propose a dual-branch framework that integrates video understanding backbones with monocular depth estimation. This hybrid representation not only improves real vs. generated content classification but also enables fine-grained categorization of generated anomalies into three distinct types, enhancing interpretability in forgery detection.  


\section{Method}
In this work, we extend the binary forgery classification to four classes, and construct the first human-centric dataset for training and evaluation. 
The detailed explanation of HFV’s construction pipeline and the generation of corresponding pseudo-labels are introduced in \cref{subsec:haf}. Also, in \cref{sub:hfr}, we describe the process of dynamically fusing features extracted from both a video understanding backbone and a monocular depth estimation model to create a robust representation of video generation anomalies. \cref{sub:loss} presents the method for enhancing the loss function with pseudo-label rankings as a confidence score.

\subsection{Problem Formulation}
Let $X \in \mathbb{R}^{T \times C \times W \times H}$ be a video input consisting of a sequence of video frames, with  human-verified generated labels $L(v)$, where $L(v) \in \{0, 1, 2, 3\}$. The values of $L(v)$ correspond to different types of anomalies: 0 for spatial anomaly, 1 for appearance anomaly, 2 for motion anomaly, and 3 for real human action videos. Our objective is to generate the final prediction $\mathit{Y}$ such that it closely approximates $L(v)$. In our work, we aim to learn a Human Forgery Representation $f_{HFR}$ that maps the input $X$ to the prediction $\mathit{Y}$. The training involves minimizing loss functions between $L(v)$ and the predicted $\mathit{Y}$. To achieve this, we require a set of videos with labels to learn the mapping from $X$ to $\mathit{Y}$. We provide such annotations in this work, and the detailed processes are described in \cref{subsec:haf}.

\begin{figure*}
  \centering
\includegraphics[width=0.9\linewidth]{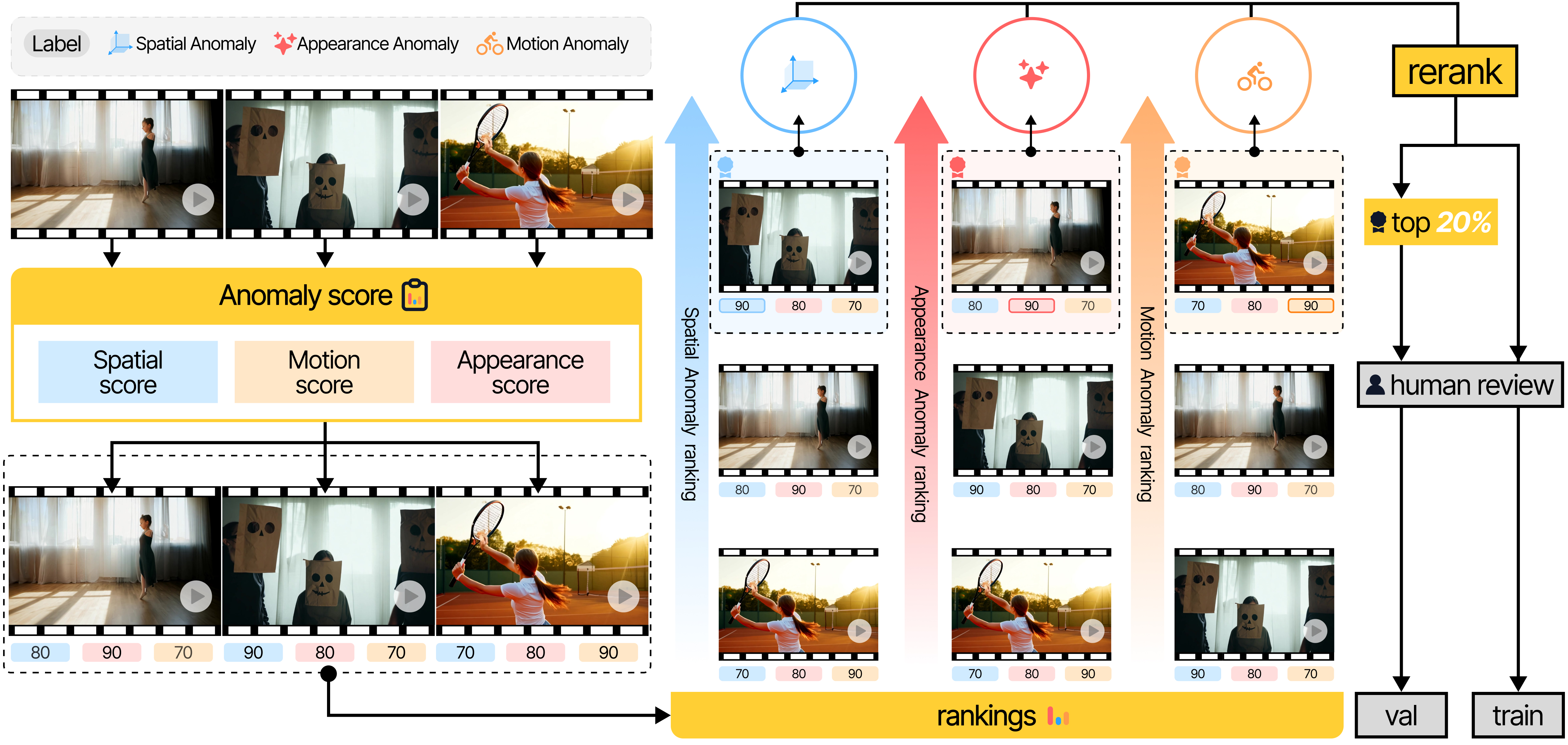}
  \caption{\textbf{The evolution process of pseudo-label generation.} Videos are processed through an anomaly scoring mechanism to produce three scores: spatial, appearance, and motion anomalies. Based on these scores, videos are ranked by anomaly type, with the highest-ranked anomaly determining the final label. During dataset construction, videos with the same anomaly label are re-ranked, and the top 20$\%$ are reviewed for strong anomalies and assigned to the validation set. The remaining 80$\%$ are also reviewed for corresponding anomalies and assigned to the training set.}
  \label{fig:label}
    \vspace{-5pt}
\end{figure*}

\subsection{Human-centric Forgery Video (HFV) Dataset}
\label{subsec:haf}

\subsubsection{Dataset Composition}
To address this new task, we constructed a comprehensive Human-centric Forgery Video (HFV) Dataset  that includes faked human action videos generated by nine state-of-the-art video generation models: MiniMax\cite{minmax}, Gen-3\cite{gen3}, Vchitect-2.0 (VEnhancer)\cite{he2024venhancer}, Kling\cite{kling}, CogVideoX-5B\cite{yang2024cogvideox}, Vchitect-2.0-2B\cite{vchitect}, CogVideoX-2B\cite{yang2024cogvideox}, Pika\cite{pika}, and Gen-2\cite{gen2}. Notably, we observed significant performance variations across models released at different times, with later models generally achieving more realistic video quality and improved overall temporal consistency. However, previous detection approaches do not account for these variations, nor do they clearly address the quality of the synthetic datasets used\cite{AIGVDet24,on-learning-multi-modal-forgery-representation-for-diffusion-generated-video-detection}.

For HFV, we selected top-ranked video generation models based on VBench benchmark evaluations\cite{huang2024vbench} and filtered out videos specifically depicting human actions from these models as synthetic human action samples. The real human action samples in HFV are sourced from the Kinetics-400 (K400) dataset\cite{kay2017kinetics}. \cref{tab:HAFF} provides an overview composition of the HFV dataset, detailing the characteristics (resolution $\geq$ 224$\times$224, frame count, duration, and frame rate) of synthesized videos from nine generation models and selected K400 videos.To ensure diversity, HFV includes approximately 20 types of backgrounds and 80 categories of human activities, balancing representativeness and the limitations of current generative models.

\subsubsection{Human Forgery Types}
Based on extensive observations and inspired by recent research findings\cite{xue2024human,chang2024matters,lei2024comprehensive}, we identified three main types of anomalies in generated human-centric videos that make simple binary classification insufficient, highlighting the need for further exploration into multi-class classification. Additionally, these three anomaly categories can be evaluated using scoring mechanisms from some existing video generation benchmarks\cite{huang2024vbench,liu2024evalcrafter} .

\begin{figure}
  \centering
\includegraphics[width=1\linewidth]{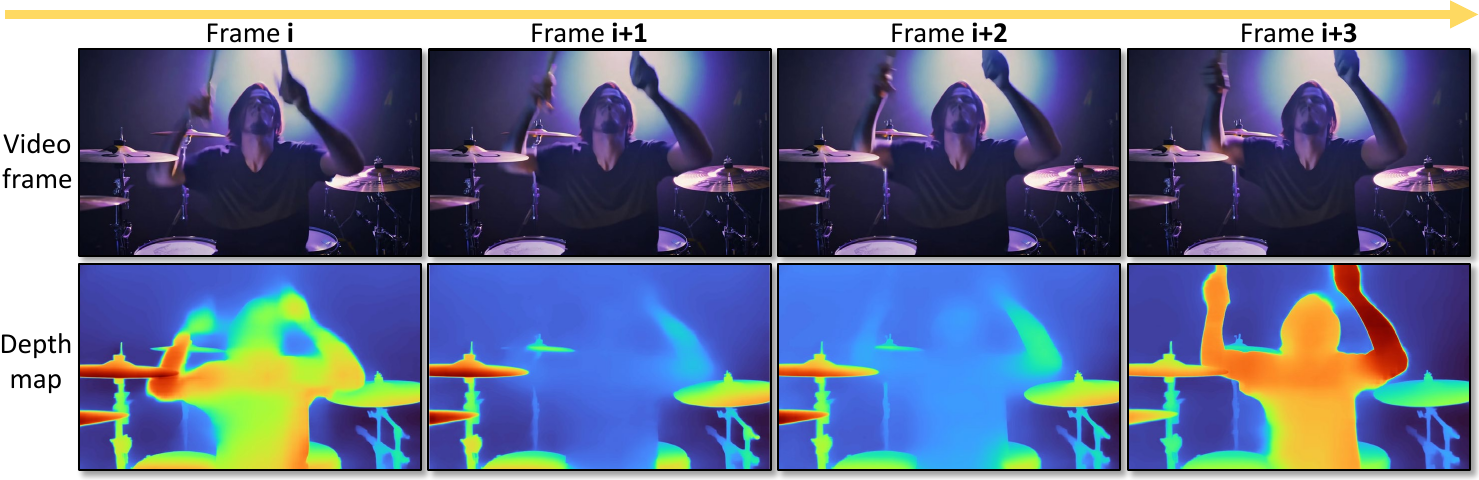}
  \caption{\textbf{Visual illustration of spatial anomaly in a generated video sequence}. Consecutive frames from a synthetic video (top) and their depth maps (bottom) reveal inconsistencies—e.g., unnatural clipping between hand and drum—despite visual appearance remaining similar.}
  \label{fig:depth}
  \vspace{-10pt}
\end{figure}


\vspace{0.2em}
\noindent\textbf{Spatial Anomaly.} This anomaly arises from incorrect spatial logic, leading to unnatural interactions such as hands passing through objects or inconsistent object scaling. To quantify spatial distortions, we utilize depth distortion errors derived from monocular depth maps, generated using the SOTA Depth Pro~\cite{bochkovskii2024depth}. Specifically, we compute distortion errors~\cite{teed2020raft, lai2018learning,lei2020blind,qi2023fatezero} by comparing optical flow maps between the depth maps of adjacent frames, leveraging RAFT~\cite{teed2020raft} for flow estimation. This process measures the deviation between predicted and observed depth, providing a robust metric for spatial anomaly detection. \cref{fig:depth} is a typical example of a spatial anomaly.

\begin{figure*}
  \centering
\includegraphics[width=0.9\linewidth]{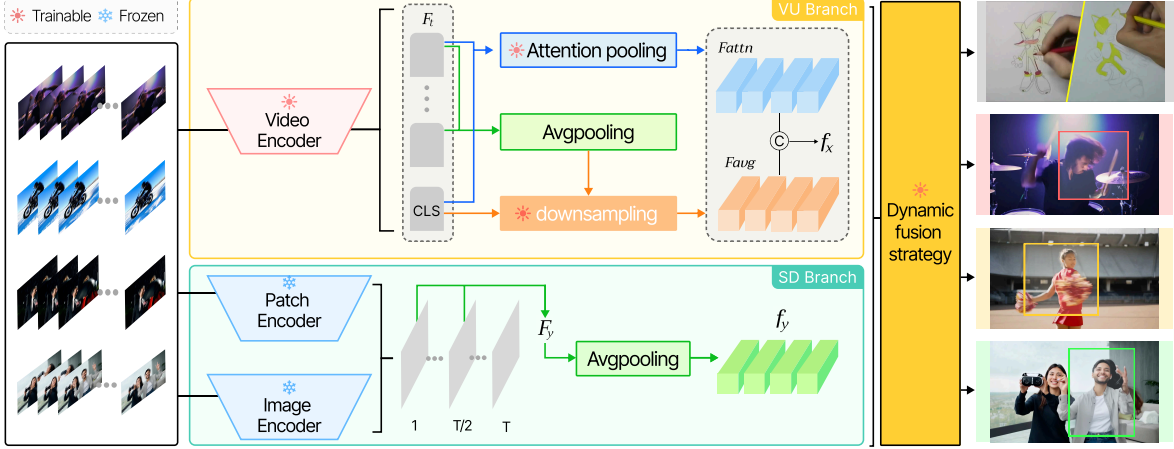}
\vspace{-5pt}
  \caption{\textbf{Dual-branch framework for human-centric video forgery detection.} The model comprises a spatio-temporal video understanding (VU) branch and a frozen spatial-depth (SD) branch. The VU branch encodes video frames using a transformer-based encoder to extract attention- and average-pooled features, capturing temporal and semantic inconsistencies. The SD branch processes frames via a depth encoder to capture spatial anomalies. Outputs from both branches are dynamically fused to detect various types of human-related forgery, as illustrated on the right.}
  \label{fig:short}
  \vspace{-10pt}
\end{figure*}


\vspace{0.2em}
\noindent\textbf{Appearance anomaly.} This anomaly involves the failure to maintain a consistent appearance of characters or objects across frames (e.g., a young girl gradually transforming into an adult woman, or scissors failing to retain a stable shape while rotating). To evaluate appearance consistency, we used a pre-trained CLIP model\cite{radford2021learning} for background consistency and DINOv2\cite{oquab2023dinov2} for subject consistency.

\vspace{0.2em}
\noindent\textbf{Motion anomaly.} This type describes unnatural or incoherent movement patterns, such as a human torso rotating a full 360 degrees. We assessed motion anomaly using a distortion error metric based on optical flow\cite{teed2020raft, lai2018learning,lei2020blind,qi2023fatezero}, employing the same approach to quantify unnatural movements.

These categories and scoring mechanisms provide a foundation for more nuanced, interpretable classification in generative video evaluation. For more details on the scoring criteria and additional examples of the three types of anomalies, please refer to the Supplementary Material.

Each video accordingly acquired three distinct anomaly scores via the scoring mechanism employed. When selecting data from a particular generation model for training and validation, we ranked the videos in descending order based on each anomaly score. Each video received rankings on three types of anomalies, with lower rankings suggesting the corresponding anomaly is less pronounced. We assigned the anomaly label of each video based on its highest-ranking anomaly type. For example, if a video’s highest ranking score was in appearance anomaly, it was labeled as an appearance anomaly.



For each video $ v_i $, its anomaly score comprises three components: spatial anomaly $ S_s(v_i) $, appearance anomaly $ S_a(v_i) $, and motion anomaly $ S_m(v_i) $. Subsequently, three corresponding anomaly ranks $ R_s(v_i) $, $ R_a(v_i) $, and $ R_m(v_i) $ are computed for each video, where a higher rank indicates greater anomaly severity.

\begin{equation}
  R(v_i) = \{ R_s(v_i), R_a(v_i), R_m(v_i) \}
  \label{eq:set_of_R}
\end{equation}

For \( v_i \), take the highest rank among \( R(v_i) \), and the actual rank with the smallest number is considered as the label for the video, denoted as \( L(v_i) \). It is defined as follows:
\begin{equation}
L(v_i) = \arg\min_{c \in \{s, a, m\}} R_c(v_i)
\end{equation}

where \( c \in \{s, a, m\} \) corresponds to spatial, appearance, and motion anomaly, respectively.

For example, if \( R_a(v_i) < R_m(v_i) \) and \( R_a(v_i) < R_s(v_i) \), then \( L(v_i) = a \), meaning the video is assigned the "appearance anomaly" label. The process of label generation for the dataset is illustrated in \cref{fig:label}. Although fewer than 2\% of videos in the HFV dataset contain all three types of anomalies, we assign the most salient one to ensure interpretability and label consistency.

\subsubsection{Dataset Split}
For each anomaly class, we again rank internally, maintaining the same ranking rules as above. The top 20$\%$ with the most obvious anomalies, supplemented by human review, is selected as the validation set, while the remaining 80$\%$, is used as the training set.

\begin{table*}[h!]
\centering

\caption{Video forgery detection performance on the HFV dataset measured by \textbf{multi-class} ACC (\%) and AUC (\%). [ACC/AUC in the Table;Key: \textbf{Best};Avg.: Average].}
\label{tab:mul}
\setlength{\tabcolsep}{4pt} 
\renewcommand{\arraystretch}{1.2} 
\newcommand{\cmark}{\checkmark} 
\scalebox{0.9}{
\begin{tabular}{cccccccccc}
\toprule

\multirow{2}{*}{Method}& \multirow{2}{*}{MiniMax} & \multirow{2}{*}{Gen-3} & Vchitect-2.0 & \multirow{2}{*}{Kling}& CogVideoX- & Vchitect- & \multirow{2}{*}{pika} & \multirow{2}{*}{Gen-2} & \multirow{2}{*}{Avg.} \\
 &  &  & (VEnhancer) & &5B & 2.0-2B &  & \\
\midrule
 CNNDet\cite{wang2020cnn}& 57.8/75.8 & 58.0/79.5 & 61.3/80.6 & 56.4/75.4 & 68.0/85.6 & 56.3/75.9 & 61.2/77.5 & 55.8/72.3&59.4/77.8 \\
 DIRE\cite{wang2023dire}& 60.8/78.2 & 59.5/80.1 & 64.7/82.0 & 59.2/77.1 & 69.6/86.9 & 56.0/77.3 & 60.4/78.1 & 53.9/72.0 &60.5/79.0\\
 F3Net\cite{qian2020thinking}& 50.5/74.8 & 56.5/66.4 & 50.5/74.9 & 48.0/74.0 & 57.0/71.3 & 50.0/76.6 & 50.0/79.6 & 43.5/74.4&50.8/74.0 \\
 Uni-FD\cite{ojha2023towards}& 55.3/78.9 & 53.5/79.1 & 64.1/84.4 & 47.6/74.7 & 72.8/89.4 & \textbf{68.3}/85.8 & 64.2/83.2 & 53.6/74.8 &59.9/81.3\\
TimeSformer\cite{bertasius2021space}& 63.3/84.1 & 65.2/84.2 & 68.4/86.0 & 60.2/82.2 & 70.8/88.0 & 62.5/82.5 & 65.3/85.1 & 55.9/78.5 &64.0/83.8\\
MM-Det\cite{on-learning-multi-modal-forgery-representation-for-diffusion-generated-video-detection}& 66.5/85.8 & 63.8/85.2 & 67.3/87.4 & 52.6/79.7 & 69.2/88.2 & 57.1/83.2 & 64.5/85.2 & 57.7/81.2 &62.3/84.5\\
 Ours & \textbf{70.4/88.2} & \textbf{73.8/88.6} & \textbf{72.3/89.5} & \textbf{65.8/87.5} & \textbf{75.6/92.1} & 66.5/\textbf{86.2} & \textbf{69.6/88.0} & \textbf{64.2/83.4} & \textbf{69.8/87.9} \\
\bottomrule
\end{tabular}
}
\end{table*}

\subsection{Human Forgery Representation}
\label{sub:hfr}

To better capture the anomalous features in video generation, we propose a novel Human Forgery Representation (HFR), as depicted in \cref{fig:short}. This representation fully exploits the advantages of large-scale visual backbone models in extracting appearance, motion consistency, and spatial depth features. HFR consists of two feature extraction branches: a Video Understanding Branch and a Spatial Depth Branch, which respectively extract spatiotemporal and spatial depth consistency features, and integrate to form a complete anomaly feature representation.

\vspace{0.2em}
\noindent\textbf{Video Understanding Branch.}
The Video Understanding Branch is based on the InternVideo2~\cite{wang2024internvideo2} model, which employs a video encoder structure combined with an attention pooling layer, capable of efficiently extracting spatiotemporal consistency features from videos. Let the input video be represented as a tensor containing $T$ frames $X \in \mathbb{R}^{T \times H \times W \times C}$, where each frame is represented as $x \in \mathbb{R}^{H \times W \times C}$.

After processing through the video encoder, the features of the input video are transformed into a spatiotemporal consistency feature matrix $F_t \in \mathbb{R}^{(T \times L) \times C}$, where $L$ denotes the encoder feature dimension. The CLS token aggregates global information from the entire feature, and we extract this token from $F_t$ while performing average pooling on the remaining features across the $T \times L$ dimensions to obtain the initial spatiotemporal feature $F_{\text{avg}}$. And the $F_t$ is processed through an attention pooling layer that generates a global query via average pooling and applies cross-attention using the original features as keys and values, resulting in the enhanced spatiotemporal consistency feature $F_{\text{attn}}$. Finally, we concatenate $F_{\text{avg}}$ with the $F_{\text{attn}}$ to obtain the output feature of the Video Understanding Branch:

\begin{equation}
    f_x = \left[ F_{\text{avg}}, F_{\text{attn}} \right]
\end{equation}

\noindent\textbf{Spatial Depth Branch.}
The Spatial Depth Branch is based on the encoder structure of the monocular depth estimation model Depth Pro\cite{bochkovskii2024depth}, which consists of a Patch Encoder and an Image Encoder, used for extracting depth consistency features from videos. Let the input video be $X \in \mathbb{R}^{T \times H \times W \times C}$, we select the first frame $X_1$ and the middle frame $X_{\frac{T}{2}}$ to input into the joint encoder. After encoding, the output feature map is $F_y \in \mathbb{R}^{2 \times 1024 \times 48 \times 48}$, and by performing average pooling on it, we obtain the depth feature vector:

\begin{equation}
    f_y = \text{AvgPool}(F_y)
\end{equation}

\noindent\textbf{Dynamic Fusion Strategy.}
To align the feature vectors of the Video Understanding Branch and the Spatial Depth Branch in feature space, we apply a linear transformation to the output $f_x$ of the Video Understanding Branch to match its dimensionality with the depth feature $f_y$. Ultimately, a weighted fusion of the two features is performed using a learnable parameter $\alpha \in [0, 1]$ to obtain the Human Forgery Representation $f_{\text{HFR}}$. This representation is then projected to the final prediction scalar value $Y$ using a linear projection layer(i.e., PROJ). The combined equation for this process is:

\begin{equation}
    Y =  \text{PROJ}\left( f_{\text{HFR}}\right)=\text{PROJ}\left( \alpha \cdot f_x + (1 - \alpha) \cdot f_y \right)
\end{equation}

Here, $\alpha$ learns the weighting ratio between features, ensuring that the fused feature $f_{\text{HFR}}$ contains both appearance, motion and spatial depth features before being projected to the scalar prediction $Y$.

\subsection{Rank-based Confidence Enhancement}
\label{sub:loss}
\textbf{Process Description.}
In this method, to guide the model to pay more attention to samples with higher confidence (i.e., those with higher rankings and more obvious anomalies), we designed a ranking-based confidence enhancement mechanism. Specifically, we normalize the ranking of each sample and use it as additional information to weight the loss. This mechanism, through a function mapping, converts ranking information into confidence weights and uses it to adjust the effect of the loss function.

\begin{table*}[h!]
\centering
\caption{Ablation analysis measured by \textbf{multi-class} ACC (\%) and AUC (\%). [ACC/AUC in the Table; Key: \textbf{Best}; Avg.: Average; ].}
\label{tab:ablation}
\setlength{\tabcolsep}{1pt} 
\renewcommand{\arraystretch}{1.2} 
\newcommand{\cmark}{\checkmark} 
\scalebox{0.9}{
\begin{tabular}{cccccccccccccc}
\toprule
 &   & Modules & & \multirow{2}{*}{MiniMax} & \multirow{2}{*}{Gen-3} & Vchitect-2.0 & \multirow{2}{*}{Kling}& CogVideoX- & Vchitect- & \multirow{2}{*}{pika} & \multirow{2}{*}{Gen-2} & \multirow{2}{*}{Avg.} \\
Backbone & Cat&Depth   &  Loss &  &  & (VEnhancer) & &5B & 2.0-2B &  & \\
\midrule
\cmark &        &       &    & 64.9/86.5 & 57.4/86.2 & 67.5/87.7 & 64.7/85.2 & 68.3/89.7 & 53.5/84.5 & 64.7/86.6 & 54.9/83.1 &63.3/87.4 \\
\cmark & \cmark &       &     & 67.3/86.7 & 67.0/86.1 & 70.0/87.7 & 59.3/85.3 & 71.8/89.7 & 60.6/84.9 & 67.3/87.1 & 61.7/82.7&65.6/86.3 \\
\cmark & \cmark & \cmark &     & 69.1/88.0 & 72.8/88.0 & 72.1/89.4 & 63.6/87.1 & 75.0/91.8  &65.8/\textbf{86.3} & \textbf{70.1/88.4} & \textbf{65.3/83.4} &69.2/87.8\\
\cmark & \cmark & \cmark & \cmark & \textbf{70.4}/\textbf{88.7} & \textbf{73.8}/\textbf{88.6} & \textbf{72.3}/\textbf{89.5} & \textbf{65.8/87.5} & \textbf{75.6/92.1} & \textbf{66.5}/86.2 & 69.6/88.0 & 64.2/\textbf{83.4} &\textbf{69.8/88.0}\\
\bottomrule
\end{tabular}
}
\vspace{-5pt}
\end{table*}

\noindent\textbf{Mathematical Formulation.}
For each sample within each anomaly class, where the loss value for each sample is \( \mathcal{L}_i \), and the ranking is \( r_i \) (a lower numeric ranking value indicates higher ranking). The following are the steps in the formulation:
Ranking Normalization:
Let \( r_i \) be the ranking of sample \( i \) and assume the rankings range from 1 to \( n \) (corresponding to the total number of samples in the anomaly class). We normalize the rankings as follows:
\begin{equation}
    \hat{r}_i = \frac{r_i}{n}
\end{equation}

where \( \hat{r}_i \in [0, 1] \), representing the normalized ranking of sample \( i \).
Confidence Coefficient Calculation:
Calculate a confidence coefficient \( \alpha_i \) using the \( \hat{r}_i \) value, where a higher \( \hat{r}_i \) value corresponds to lower confidence (higher penalty weight). We map the ranking using the function \( \log(e + x) \):
\begin{equation}
    \alpha_i = \log(e + \hat{r}_i)
\end{equation}

Thus, when \( \hat{r}_i \) is close to 0, \( \alpha_i \) is close to 1; when \( \hat{r}_i \) is close to 1, \( \alpha_i \) is relatively larger.
Weighted Loss Function:
Apply \( \alpha_i \) to the loss \( \mathcal{L}_i \) of each sample to obtain the weighted loss \( \mathcal{L}_i^{\text{weighted}} \):
\begin{equation}
    \mathcal{L}_i^{\text{weighted}} = \alpha_i \cdot \mathcal{L}_i = \log(e + \hat{r}_i) \cdot \mathcal{L}_i
\end{equation}

Total Loss:
The final total loss is the average of the weighted losses of all samples in the batch (\( N \) is the total number of samples in the batch):
\begin{equation}
    \mathcal{L}_{\text{total}} = \frac{1}{N} \sum_{i=1}^{N} \mathcal{L}_i^{\text{weighted}} = \frac{1}{N} \sum_{i=1}^{N} \log(e + \hat{r}_i) \cdot \mathcal{L}_i
\end{equation}

\noindent\textbf{Explanation.}
This ranking-based confidence enhancement mechanism adjusts the loss weights so that the model pays more attention to samples with with higher rankings during the training process, as these samples have higher confidence. Therefore, this method effectively guides the model to prioritize learning the features of high-confidence samples, thereby improving the ACC of the model.

\section{Experiments}

\begin{table*}[h!]
\centering
\caption{Comparison of \textbf{general binary classification} on \textbf{HFV} measured by ACC (\%) and AUC (\%).[ACC/AUC in the Table; Key: \textbf{Best};  Avg.: Average;*: the use of the released best model, as its training code is not publicly available.].}
\label{tab:general}
\setlength{\tabcolsep}{4pt} 
\renewcommand{\arraystretch}{1.2} 
\newcommand{\cmark}{\checkmark} 
\scalebox{0.9}{
\begin{tabular}{cccccccccc}
\toprule

\multirow{2}{*}{Method}& \multirow{2}{*}{MiniMax} & \multirow{2}{*}{Gen-3} & Vchitect-2.0 & \multirow{2}{*}{Kling}& CogVideoX- & Vchitect- & \multirow{2}{*}{pika} & \multirow{2}{*}{Gen-2} & \multirow{2}{*}{Avg.} \\
 &  &  & (VEnhancer) & &5B & 2.0-2B &  & \\
\midrule
  CNNDet\cite{wang2020cnn} & 87.5/89.3 & 87.6/89.2 & 87.6/89.3 & 80.9/81.7 & 87.7/89.6 & 87.2/89.0 & 87.5/89.3 & 87.5/89.3 & 86.7/88.3 \\
DIRE\cite{wang2023dire}& 87.9/89.6 & 88.4/90.0 & 88.1/89.6 & 82.6/83.4 & 88.3/90.1 & 87.8/89.4 & 88.3/90.0 & 87.9/89.6 & 87.7/89.0 \\
 F3Net\cite{qian2020thinking}& 88.7/90.0 & 82.0/87.0 & 84.0/87.0 & 77.7/74.5 & 86.3/90.5 & 82.7/86.0 & 93.0/95.5 & 81.0/79.5 &  84.4/86.3\\
 Uni-FD\cite{ojha2023towards}& 83.8/97.6 & 87.3/99.8 & 70.3/97.8 & 50.1/73.4 & 88.3/98.8 & 78.5/96.3 & 90.2/98.7 & 98.5/99.8 & 80.9/95.3 \\
 TimeSformer\cite{bertasius2021space}& 96.1/99.5 & 95.9/99.3 & 96.3/99.6 & 88.0/95.1 & 96.5/99.7 & 96.0/99.3 & 96.5/99.7 & 96.4/99.7 &95.2/97.9\\
 HiFi-Net*\cite{guo2023hierarchical}& 57.9/54.6 & 57.0/51.8 & 57.1/58.7 & 57.6/36.7 & 57.9/48.7 & 57.8/46.1 & 57.8/59.5 & 57.8/59.5&57.6/52.0 \\
 MM-Det\cite{on-learning-multi-modal-forgery-representation-for-diffusion-generated-video-detection}&97.0/99.8 & \textbf{99.4/100} & \textbf{98.6/99.9} & 85.7/97.9 & 99.2/\textbf{100} & \textbf{99.3/100} & 97.5/99.8 &\textbf{99.7/100} & 97.0/99.7 \\
Ours& \textbf{99.0/100} & 97.4/99.9 & 98.1/\textbf{99.9} & \textbf{90.3/99.3} & \textbf{99.4/100} & \textbf{99.3/100} & \textbf{99.4/100} & 99.4/\textbf{100} &\textbf{97.8/99.9}\\

\bottomrule
\end{tabular}
}
\vspace{-5pt}
\end{table*}

\subsection{Experiments Setup}
In the experiments, we use the HFV dataset for evaluation. In training, we select a total of 1000 videos, including 720 human-centric forgery videos from CogVideoX-2B\cite{yang2024cogvideox}, consisting of 221 with appearance anomalies, 224 with spatial anomalies, and 275 with motion anomalies. Additionally, we include 280 real human action videos from the K400 dataset\cite{kay2017kinetics} to form the training set. For each of the three anomaly categories, we select the top 20$\%$ of videos based on anomaly scoring and conduct a human review to confirm the prominence of anomalies. These samples are used to form the validation set. For real videos, we randomly select 20$\%$ for validation, with the remainder used for training. Furthermore, we sample 530 real videos from the K400 dataset and combine them with forgery videos from eight other generation models to create eight evaluation datasets. More training details can be found in the Supplementary Material.


To ensure a fair comparison, we benchmark our method against 6 recent detection approaches. CNN-Det~\cite{wang2020cnn} employs a CNN classifier for forgery detection, while F3Net~\cite{qian2020thinking} leverages frequency-domain features. DIRE~\cite{wang2023dire} utilizes DDIM-based~\cite{song2020denoising} reconstruction to detect diffusion-generated images, and Uni-FD~\cite{ojha2023towards} exploits CLIP~\cite{radford2021learning} feature space for classification. For video-level detection, TimeSformer~\cite{bertasius2021space} models spatiotemporal relationships via self-attention, whereas MM-Det~\cite{on-learning-multi-modal-forgery-representation-for-diffusion-generated-video-detection} extracts multimodal features via MLLM and reconstructs content with VQVAE. We evaluate performance using multi-class ACC and AUC.

\subsection{Comparison to Existing Detectors}
To upgrade existing detectors into multi-class detectors, we adopted the following modification strategy: For detectors equipped with a linear layer, we adjusted the number of output ports in the linear layer to 4. Furthermore, we standardized the loss function for all modified detectors to the multi-class cross-entropy loss.

As shown in \cref{tab:mul}, In the multi-classification task, our proposed HumanSAM achieves SOTA performance in detecting human-centered forgery videos. On average, it surpasses the second-best method, TimeSformer~\cite{bertasius2021space}, by 5.8$\%$ in ACC and MM-Det~\cite{on-learning-multi-modal-forgery-representation-for-diffusion-generated-video-detection} by 3.4$\%$ in AUC. Specifically, prior methods based on pre-trained CLIP features, such as Uni-FD\cite{ojha2023towards}, perform well on certain types of diffusion-generated content (e.g., CogVideoX-5B, Vchitect-2.0-2B). However, they struggle with video generation models that produce high-quality, temporally and spatially consistent results (e.g., MiniMax\cite{minmax}, Gen-3\cite{gen3}).                                                                                                                                                                                       

F3Net\cite{qian2020thinking} performs the worst in this multi-classification experiment, indicating that relying solely on frequency domain information for multi-classification is challenging. DIRE\cite{wang2023dire}, which uses diffusion model reconstruction for classification, shows stable performance across most diffusion models but fails to achieve superior results. CNNDet\cite{wang2020cnn}, a simpler classifier, achieves results comparable to CLIP-based methods after fine-tuning, underscoring the importance of the dataset we proposed.


Overall, our approach leverages representations that combine spatiotemporal consistency features from video foundation models and spatial depth features from monocular depth estimation, yielding stronger performance and greater robustness. On the HFV dataset, our method consistently achieves the best results across all metrics.

Experimental results demonstrate that our proposed HFR is more robust. It not only outperforms detection methods that rely solely on frame-level features but also surpasses approaches that simply combine appearance and motion features or depend only on spatiotemporal information from video frame sequences. This is because HFR more deeply exploits the representational potential of video content.

\subsection{Ablation Study}




Based on the ablation analysis in \cref{tab:ablation}, each component of our method progressively enhances performance. Using only the InternVideo2~\cite{wang2024internvideo2} backbone provides a low baseline, achieving an average ACC of 63.3$\%$  and AUC of 87.4$\%$ . Integrating the Cat module, which fuses $F_{\text{attn}}$ and $F_{\text{avg}}$ features, improves feature representation, raising ACC to 65.6$\%$  and AUC to 86.3$\%$ . Incorporating the depth module~\cite{bochkovskii2024depth} further boosts classification, particularly on Kling and CogVideoX-5B datasets, reaching 69.2$\%$ ACC and 87.8$\%$ AUC. Finally, adding the rank-based confidence loss refines classification confidence, yielding the highest overall performance at 69.8$\%$ ACC and 88.0$\%$ AUC. These results demonstrate that each module contributes to improving both accuracy and robustness across different datasets.

\subsection{Experiments on general binary classification}
To validate the effectiveness of our framework, we conducted general binary classification experiments (real vs. fake) while maintaining SOTA performance. Notably, our model achieves 90.3\% ACC and 99.3\% AUC even on the challenging Kling dataset. TimeSformer\cite{bertasius2021space}, a self-attention-based video understanding model, effectively captures both appearance and motion features, achieving strong performance across datasets. Meanwhile, MM-Det\cite{on-learning-multi-modal-forgery-representation-for-diffusion-generated-video-detection} enhances spatial artifacts via VQVAE reconstruction, excelling on Gen\cite{gen3,gen2} and Vchitect-2.0\cite{vchitect,he2024venhancer} series datasets, where appearance fidelity is the primary objective.  
Additionally, we performed a label-mapping experiment by merging the three anomaly categories, observing that underperforming methods benefit from this reconfiguration, leading to improved binary classification performance. See ~\cref{tab:general} and  ~\cref{tab:HAFF} in the supplement for detailed results. Extensive quantitative analyses, including confusion matrices, F1-scores, attention-based localization, and cross-dataset generalization, are provided in the supplement.

\subsection{Robustness Analysis}
\begin{table}[h]
\centering
\vspace{-8pt}
\caption{Performance of Ours on common post-processing operations measured by AUC(\%).}
\label{tab:robust-performance}
\setlength{\tabcolsep}{5pt} 
\begin{tabular}{ccccc}
\hline
N/A & Blur $\sigma=3$ & JPEG $Q=50$ & Resize $0.7$ & Mixed \\
\hline
92.1 & 88.4 & 91.8 & 91.9 & 88.2 \\
\end{tabular}
\end{table}
\vspace{-5pt}
To analyze the robustness of our method, we conduct common post-processing operations on CogVideoX-5B dataset, including Gaussian Blur (\textbf{B}) with $\sigma$=3, JPEG Compression (\textbf{C}) with $Q$=$90$, Resize (\textbf{R}) with ratio=$0.7$ and a Mixture (\textbf{M}) of all operations. As reported in \cref{tab:robust-performance}, the AUC of fine-grained classification meets a degradation
of 3.7\%(\textbf{B}), 0.3\%(\textbf{C}) 0.2\%(\textbf{R}) and 3.9\%(\textbf{M}), with all AUC above 88.2\%, indicating the robustness against unseen perturbations.
\section{Conclusion}
In this work, we extend the binary video forgery classification task to multiple classes. To do this, we propose HumanSAM that can categorize human-centric video forgeries into three primary types: human appearance, motion, and spatial anomalies. Technical designs including the dual-branch fusion and the rank-based confidence enhancement strategy are proposed for better and robust performance. We also construct the first benchmark comprising these four types of videos to facilitate research on this task. Extensive experiments demonstrate that our method exhibits superior accuracy and robustness both in binary and multi-class classification tasks, underscoring its significant value in interpretable and fine-grained forgery video detection. 
More discussions and limitations are included in the supplement.

{
    \small
    \bibliographystyle{ieeenat_fullname}
    \bibliography{main}
}
\maketitlesupplementary
\appendix

In this supplementary material, we offer further details on HumanSAM. \cref{sec:score} delves into the calculation of the three anomaly scoring mechanisms. \cref{sec:implementation} details the specifics of the experimental setup. \cref{sec:further} conducts a deeper analysis of the quantitative experimental results. \cref{add} provides additional quantitative analyses to support the effectiveness of HumanSAM. \cref{st} presents analyses of spatio-temporal information. \cref{sec:exploratory} is dedicated to exploratory experimental analyses, focusing on the outcomes of training with various forgery data sources. \cref{sec:lim} is limitations.

Additionally, we have included videos featuring human-centric anomalies as part of this supplementary material to demonstrate examples of each anomaly type.
\section{Anomaly Scoring Mechanism}
\label{sec:score}
\subsection{Spatial anomaly}
We used Depth pro\cite{bochkovskii2024depth} to generate depth maps for each frame of the video. Subsequently, we employed a technique based on optical flow distortion error\cite{lai2018learning,lei2020blind,qi2023fatezero} to quantitatively evaluate these depth maps. This technique measures the consistency of motion by monitoring the trajectory of pixel movement. In the depth maps, the pixel values represent the spatial depth of the scene.By calculating the distortion error, we are able to assess the coherence between depth maps, thereby quantifying anomalies in spatial depth.
The warping error is computed as follows:

\textbf{Optical Flow Estimation:}
For two consecutive frames $I_t$ and $I_{t+1}$, the optical flow $F_{t \to t+1}$ from frame $t$ to frame $t+1$ is obtained using a optical flow estimation network \cite{teed2020raft}.
    
\textbf{Image Warping:}
Using the optical flow $F_{t \to t+1}$, frame $I_t$ is warped to the coordinates of frame $t+1$, resulting in the warped image $\hat{I}_{t+1}$:
\begin{equation}
\hat{I}_{t+1} = W(I_t, F_{t \to t+1}),
\end{equation}
where $W(\cdot, \cdot)$ represents the warping operation based on the optical flow.
    
\textbf{Pixel-wise Difference Calculation:}
The pixel-wise difference between the warped image $\hat{I}_{t+1}$ and the predicted image $I_{t+1}$ is computed using the $L_2$ norm:
\begin{equation}
E_t = \| \hat{I}_{t+1} - I_{t+1} \|_2^2.
\end{equation}
    
\textbf{Final Score:}
The warping error $E_{\text{warp}}$ is calculated as the average of the pixel-wise differences over all consecutive frame pairs:
\begin{equation}
E_{\text{warp}} = \frac{1}{T-1} \sum_{t=1}^{T-1} E_t,
\end{equation}
where $T$ denotes the total number of frames.For specific examples, please refer to \cref{fig:spatital}.

\begin{table*}[h!]
\centering
\caption{Video forgery detection performance on the HFV dataset measured by \textbf{mapped binary classification} ACC (\%) and AUC (\%). [ACC/AUC in the Table; Key: \textbf{Best}; Avg.: Average].}
\label{tab:mapbinary}
\setlength{\tabcolsep}{4pt} 
\renewcommand{\arraystretch}{1.2} 
\newcommand{\cmark}{\checkmark} 
\scalebox{0.85}{
\begin{tabular}{cccccccccc}
\toprule

\multirow{2}{*}{Method}& \multirow{2}{*}{MiniMax} & \multirow{2}{*}{Gen-3} & Vchitect-2.0 & \multirow{2}{*}{Kling}& CogVideoX- & Vchitect- & \multirow{2}{*}{pika} & \multirow{2}{*}{Gen-2} & \multirow{2}{*}{Avg.} \\
 &  &  & (VEnhancer) & &5B & 2.0-2B &  & \\
\midrule
  CNNDet\cite{wang2020cnn}& 93.7/93.0 & 93.8/93.2 & 94.1/92.8 & 77.6/79.1 & 93.0/92.4 & 89.6/89.5 & 92.3/91.8 & 93.8/93.1 & 91.0/90.6\textbf{(+4.3/+2.3)} \\
  DIRE\cite{wang2023dire}& 92.9/92.3 & 94.0/93.3 & 93.4/92.8 & 83.1/83.9 & 93.9/93.2 & 90.6/90.3 & 93.4/92.8 & 93.1/92.5&91.8/91.4\textbf{(+4.1/+2.4)} \\
 F3Net\cite{qian2020thinking}& 92.0/90.7 & 92.0/90.7 & 88.0/84.0 & 83.5/85.0 & 91.0/89.3 & 89.5/85.7 & 91.0/88.7 & 89.5/89.7&89.6/88.0\textbf{(+5.2/+1.7)} \\
Uni-FD\cite{ojha2023towards} & 95.7/99.5 & 97.1/99.4 & 93.5/99.0 & 80.2/90.7 & 91.9/98.9 &  91.4/98.5& 93.4/99.4 & 94.9/99.6& 92.3/98.0\textbf{(+11.4/+2.7)}\\
TimeSformer\cite{bertasius2021space}& 95.6/99.6 & 95.6/99.4 & 96.0/99.7 & 87.6/95.8 & 96.0/99.7 & 95.5/99.4 & 96.1/99.8 & 96.0/99.7 & 94.8/99.1(-0.4/\textbf{+1.2})\\
MM-Det\cite{on-learning-multi-modal-forgery-representation-for-diffusion-generated-video-detection}& 98.1/99.8 & \textbf{99.1/100}& \textbf{98.7/}99.9 & 73.4/94.6 & 99.0/99.9 & 98.7/99.9 & 99.0/99.9 & 99.0/100 &95.6/99.3(-1.4/-0.4)\\
 Ours & \textbf{99.1/100} & 97.9/99.2 & 98.0/\textbf{100} & \textbf{90.0/99.6} & \textbf{99.3/100} & \textbf{99.3/100} & \textbf{99.7/100} & \textbf{99.4/100} &\textbf{97.8/99.9}(0/0)\\
\bottomrule
\end{tabular}
}
\end{table*}

\subsection{Appearance anomaly}
We referred to the methods used in VBench\cite{huang2024vbench} for calculating subject consistency and background consistency. Both calculations employ the same formula, which involves calculating the sum of the cosine similarities of image features between consecutive frames, as well as the sum of the cosine similarities between the first frame’s image features and each subsequent frame, and then averaging these total similarity scores to determine the average consistency across the frames. For subject consistency, they utilized DINO\cite{caron2021emerging}, while for background consistency, they employed CLIP\cite{lai2018learning}. However, we found that this approach becomes ineffective when dealing with scene transitions.

\begin{figure*}
  \centering
\includegraphics[width=0.95\linewidth]{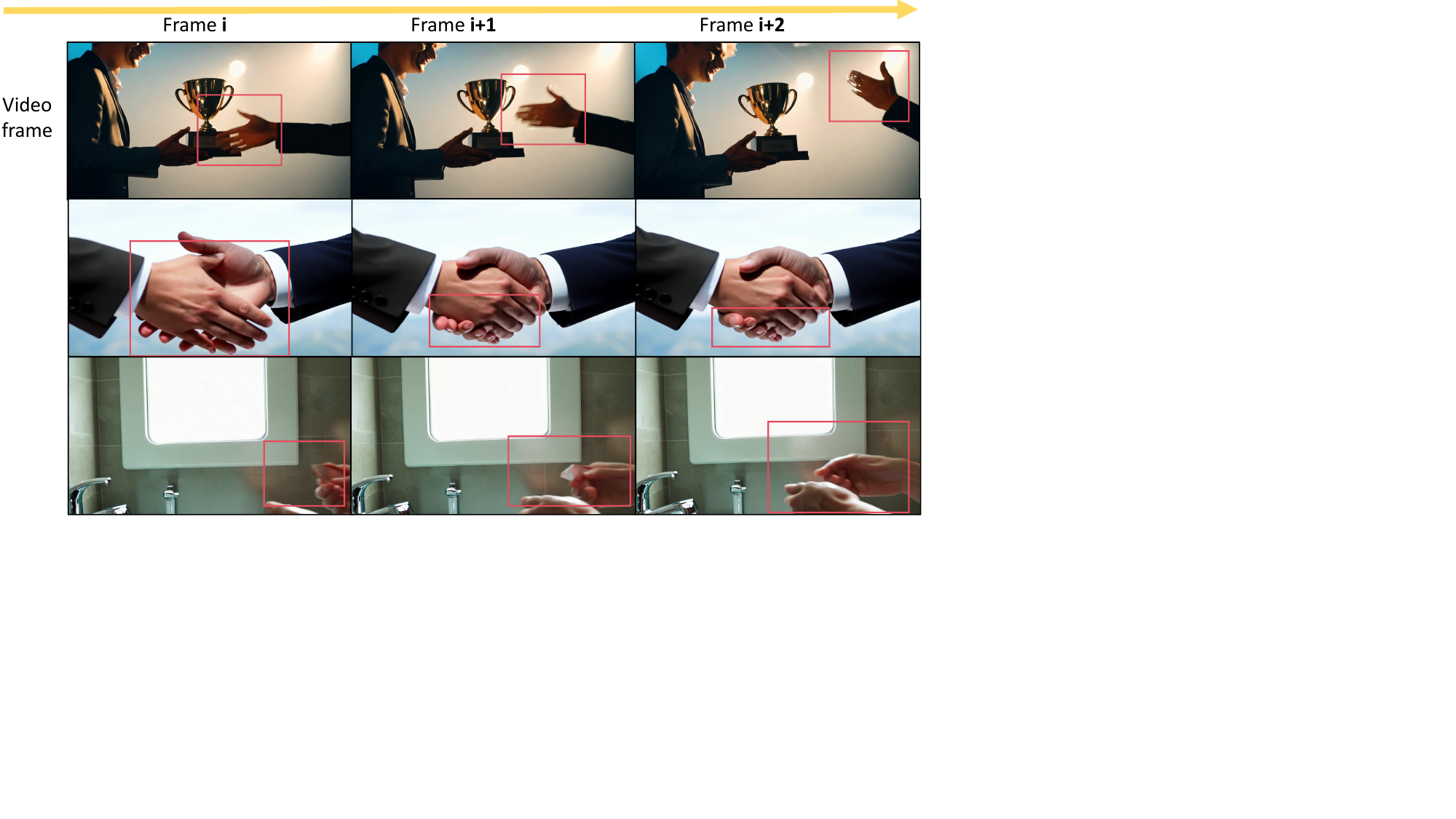}
  \caption{Some examples of appearance anomalies.It can be seen that in the first row, the hand on the right suddenly changes from an apparently left hand to a right hand. In the second row, the number of fingers on the right hand changes from six to five. In the third row, the object held in the hand gradually disappears. Generally speaking, the consistency in appearance cannot be maintained.}
  \label{fig:appearance}
\end{figure*}

To address this limitation, we abandoned the similarity calculation between the first frame and the subsequent frames and instead adopted a sliding window consistency approach. This method calculates the average similarity within a specified window, such as over a span of 5 frames. The specific calculation formula is as follows:
\begin{equation}
S_{\text{score}} = \alpha \cdot \frac{1}{T-1} \sum_{t=2}^{T} (f_{t-1} \cdot f_{t}) + \beta \cdot \frac{1}{N} \sum_{k=1}^{N} S_{\text{window},k}
\end{equation}
where $f_i$ represents the $i^{th}$ frame, the $\langle\cdot\rangle$ operation denotes the calculation of the cosine similarity of image features, and $\alpha$ and $\beta$ are the weights for the two terms, both of which are set to 0.5. The calculation formula for $S_{\text{window},k}$ is as follows:

\begin{equation}
S_{\text{window},k} = \frac{1}{W-1} \sum_{j=2}^{W} (f_{j-1} \cdot f_{j})
\end{equation}
where $W$ denotes the window size, it is specified as 5.We finally calculate the score of a video using CLIP and DINOv2\cite{oquab2023dinov2} respectively, and then take the average as the appearance anomaly score for that video.For specific examples, please refer to \cref{fig:appearance}.

\subsection{Motion anomaly}
\begin{figure*}
  \centering
\includegraphics[width=0.95\linewidth]{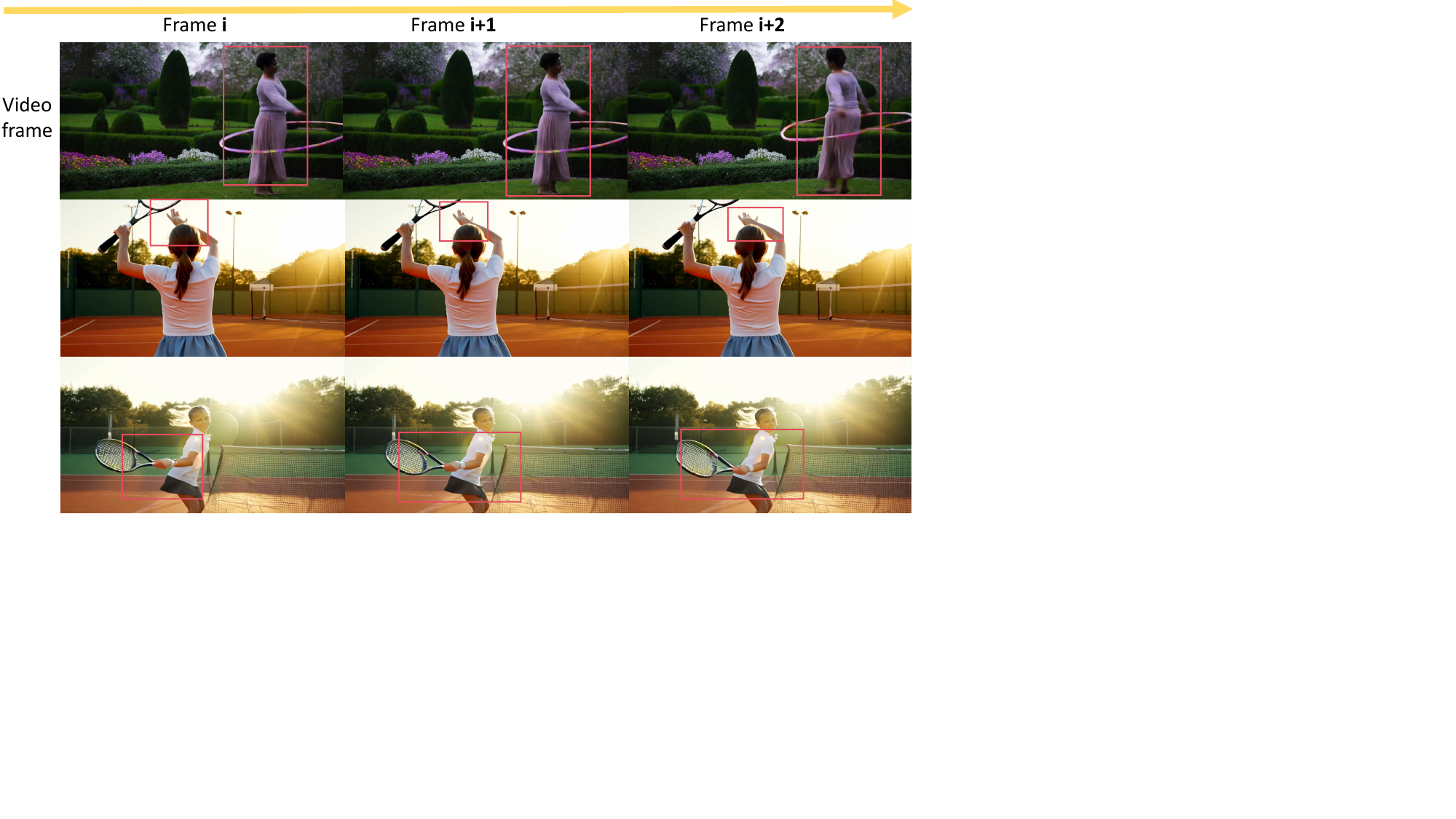}
  \caption{Some examples of motion anomalies.It can be observed that in the first row, the woman's body maintains a forward-leaning tendency, but her head suddenly rotates 180 degrees. In the second row, the girl's right hand takes on the shape of a left hand. In the third row, the girl's right hand assumes the posture of a left hand, which would be appropriate if the girl's body were rotated around. Generally speaking, the motion of the characters does not conform to normal biological motion patterns.}
  \label{fig:motion}
\end{figure*}

The calculation method for motion anomaly is to directly compute the distortion error of the video frame images, which is the same method as the distortion error calculation for spatial anomalies.For specific examples, please refer to \cref{fig:motion}.

\begin{figure*}
\centering
    \begin{subfigure}[b]{0.49\textwidth}
        \centering
        \includegraphics[width=\textwidth]{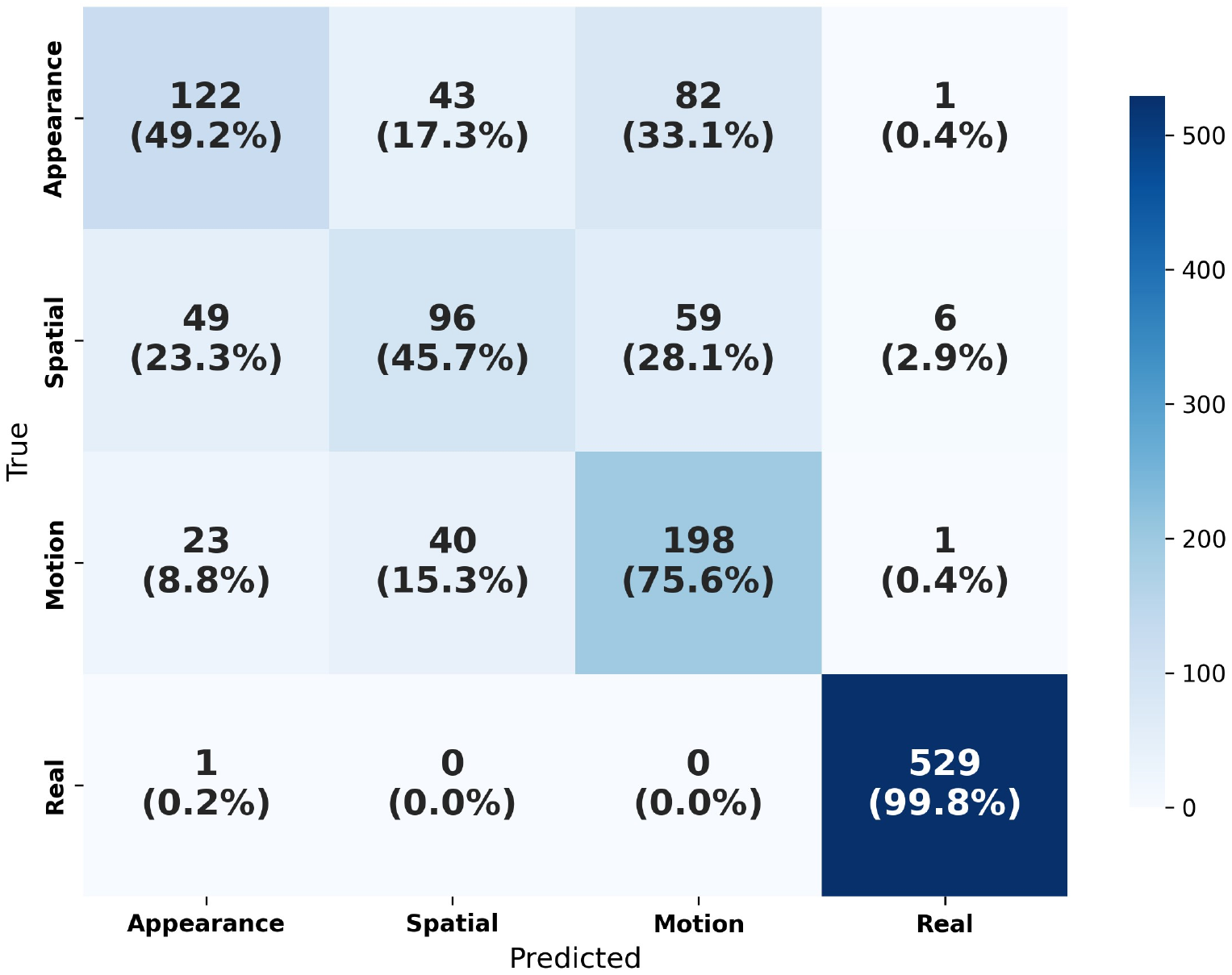}
        \caption{}
        \label{fig:four_class_cm}
    \end{subfigure}
    \hfill
    \begin{subfigure}[b]{0.49\textwidth}
        \centering
        \includegraphics[width=\textwidth]{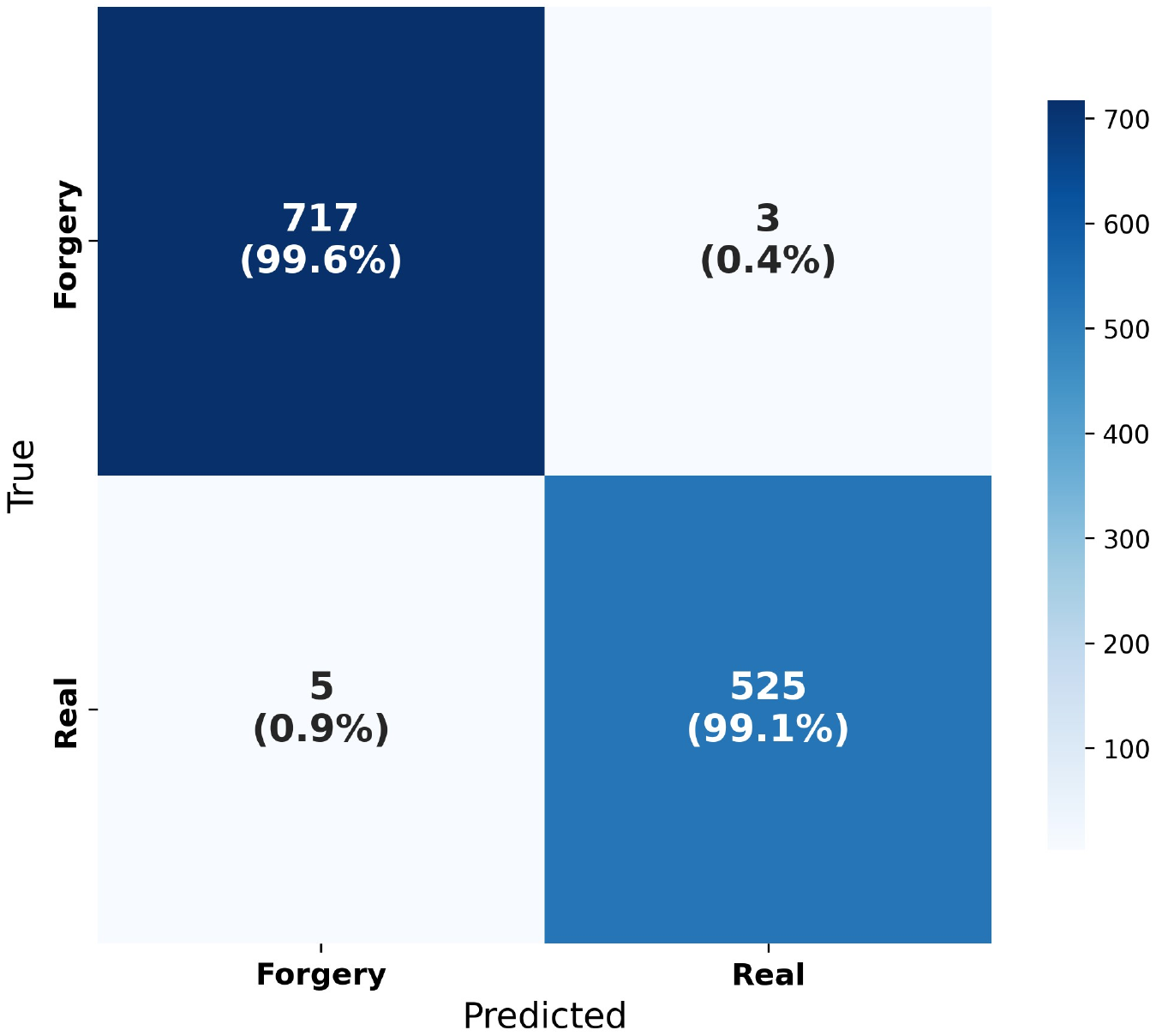}
        \caption{}
        \label{fig:binary_cm}
    \end{subfigure}
    \caption{Confusion matrices on CogVideoX-5B dataset: (a) Multi-class, (b) Binary.}
    \label{fig:confusion_matrices}
  \label{fig:spatital}
\end{figure*}

\section{Implementation Details}
\label{sec:implementation}
\subsection{Dataset}
The dataset is organized according to the nine types of forged data sources shown in main text. The order from top to bottom also corresponds to the ranking by the VBench team\cite{huang2024vbench}. This means that MinMax ranks first in the HFV, followed by Gen3, then Vchitect-2.0 (VEnhancer), with Gen-2 being the last.
\subsection{Hyperparameters of HumanSAM}
We train all parameters of the video understanding branch while freezing the parameters of the spatial depth branch. The video understanding branch selected is the distilled L version of the InternVideo2\cite{wang2024internvideo2} single modality, with a patch size of $14 \times 14$. We choose the image encoder and patch encoder of Depth pro\cite{bochkovskii2024depth} as the spatial depth branch. The final output of the video understanding branch is $f_x \in \mathbb{R}^{2816}$. The vector from the spatial depth branch, after pooling and other operations, becomes $f_y \in \mathbb{R}^{1024}$. Therefore, $f_x$ is passed through a linear layer to reduce its dimensionality to 1024. $f_x$ and $f_y$ are then combined through a trainable parameter $\alpha$ to form the final HFR.
\subsection{Training and Inference}
For the experimental resources used in training and inference, all experiments were conducted using a single NVIDIA RTX 3090 GPU with a maximum of 256G of memory.

During training, for each video, we performed segmented sampling, collecting a total of eight frames, which were then cropped to 224x224 as input.We used the AdamW optimizer with a learning rate of 2e-5 and ran for 100 epochs, selecting the best performance on the validation data from the training set.

For inference, we evaluated all models at the video level. For frame-level baselines, the final result was the average of all frame results. For video-level baselines, the results were obtained following their respective  default frame sampling and evaluation settings.


\begin{figure*}
  \centering
\includegraphics[width=1.0\textwidth]{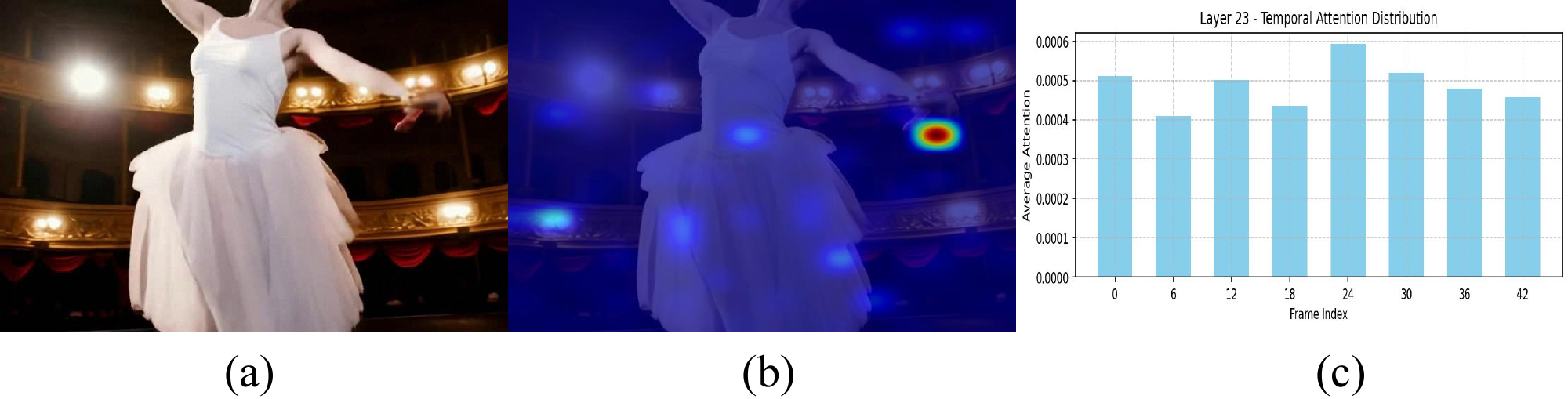}
    \caption{Spatial and temporal activation visualization on a CogVideoX-5B video. From the last Transformer layer of the video branch: (a) original frame at index 24; (b) spatial activation map of that frame; (c) temporal activation bar plot averaged over 8 sampled frames.}
    \label{fig:attention_vis}
\end{figure*}

\section{Mapped Binary Classification  Experiment}
\label{sec:further}
Due to space constraints in the main text, we supplement here the general binary classification experiments for TimeSformer and HFR. A comparison of Tab. 4 of main text and \cref{tab:mapbinary} reveals that methods with lower binary classification accuracy, such as CNNDet\cite{wang2020cnn}, DIRE\cite{wang2023dire}, F3Net\cite{tan2024rethinking} and Uni-FD\cite{ojha2023towards}, can significantly improve their binary classification performance when trained using our proposed multi-class task. However, for models like TimeSformer\cite{bertasius2021space}, MM-Det\cite{on-learning-multi-modal-forgery-representation-for-diffusion-generated-video-detection} and ours, which already achieve high accuracy in binary classification, training with the new task has little impact on their binary classification performance.

Notably, HFR achieves an average ACC of 97.8\% and an average AUC of 99.9\%, further demonstrating that our method more effectively models video features, enabling it to distinguish between real and synthetic videos with greater precision.

\section{Additional Quantitative Analyses}
\label{add}
\subsection{Confusion Matrices and Per-Class Performance}
\cref{fig:confusion_matrices} shows the confusion matrices for both fine-grained (four-class) and binary classification. Our method achieves F1-scores of 0.5508 (appearance anomaly), 0.4936 (spatial anomaly), 0.6589 (motion anomaly), and 0.9916 (real). Most confusions occur between motion and spatial classes, partially due to the motion-sensitive video branch and the limitations of the frozen depth encoder. Nevertheless, the binary classification performance remains strong and stable.

\subsection{Generalization on External Datasets}
To evaluate generalizability beyond the HFV dataset, we test on the Sora dataset, which includes a broader distribution of scenarios. HumanSAM achieves 95.3\% accuracy and 99.5\% AUC, outperforming MM-Det\cite{on-learning-multi-modal-forgery-representation-for-diffusion-generated-video-detection} (81.0\% / 98.4\%). This demonstrates the framework’s strong transferability even in the presence of diverse, non-human-specific generative content. Future work will explore methods with improved generalization across open-world forgery distributions.

\section{Spatio-Temporal Information Analyses}
\label{st}
\cref{fig:attention_vis} illustrates how our model leverages spatio-temporal cues to localize anomalies. Specifically, spatial attention highlights the girl's \textbf{distorted} right hand in frame 24, while the highest temporal score is also assigned to this frame. This demonstrates the model’s potential for precise frame-level anomaly localization. Additionally, lighting inconsistencies—though not strictly human-centric—are considered part of the appearance anomaly when they impact human actions (e.g., uneven illumination on the left side in \cref{fig:attention_vis}).

\begin{table*}[h!]
\centering
\caption{Comparison of \textbf{multi-class training} using \textbf{different }forgery video sources in the \textbf{HFV} dataset measured by ACC (\%) and AUC (\%).[ACC/AUC in the Table; Key: \textbf{Best};  Avg.: Average].}
\label{tab:exp}
\setlength{\tabcolsep}{3pt} 
\renewcommand{\arraystretch}{1.2} 
\newcommand{\cmark}{\checkmark} 
\scalebox{0.9}{
\begin{tabular}{cccccccccc}
\toprule

    Video& \multirow{2}{*}{MiniMax} & \multirow{2}{*}{Gen-3} & Vchitect-2.0 & \multirow{2}{*}{Kling}& CogVideoX- & Vchitect- & \multirow{2}{*}{pika} & \multirow{2}{*}{Gen-2} & \multirow{2}{*}{Avg.} \\
 Source&  &  & (VEnhancer) & &5B & 2.0-2B &  & \\
\midrule
  CogVideoX-2B & \textbf{70.4/88.2} & \textbf{73.8/88.6} & \textbf{72.3/89.5} & 65.8/87.5 & \textbf{75.6/92.1} & 66.5/\textbf{86.2} & \textbf{69.6/88.0} & \textbf{64.2/83.4} & \textbf{69.8/87.9}  \\
MinMax& - & 63.7/82.2 & 68.8/87.1 & \textbf{84.5}/\textbf{96.4} & 65.4/86.5 & 60.2/83.0 & 67.6 /85.1 & 55.6/80.6 & 66.5/85.6 \\

 Kling& 69.5/87.2 & 63.7/82.2 & 68.8/87.1 & - & 64.2/83.5 & 60.2/83.0 & 67.0/85.1 & 55.6/80.6 &  64.1/84.1\\
\bottomrule
\end{tabular}
}
\end{table*}

\section{Exploratory Experimental Analysis}
\label{sec:exploratory}
Due to the length constraints of the main text, some details of exploratory experiments are presented here. As shown in Tab. 1 of main text, the higher a synthetic data source ranks, the better its overall performance on the VBench benchmark\cite{huang2024vbench}. MinMax\cite{minmax} ranks first, while Kling\cite{kling} ranks fourth. As shown in \cref{tab:mapbinary}, all methods experience a sudden performance drop on Kling. To further investigate this, while keeping the original experimental settings unchanged, we replaced the CogVideoX-2B forgery data in the training set with Kling and MinMax for four-class training. This adjustment was made to compare the results across the remaining seven forgery video sources.

As shown in \cref{tab:exp}, after replacing the training data with MinMax and Kling, the overall results still demonstrate that the seventh-ranked CogVideoX-2B achieves the best performance. Upon closer examination of the CogVideoX-2B videos, we observed that they still exhibit noticeable gaps compared to realistic human behavior. This leads us to hypothesize that lower-ranked synthetic data sources may contain more of the three types of anomalies, which in turn benefits the proposed HFR in learning anomalous features.
Kling and MinMax share identical metrics across most forgery data sources, indicating that their anomalous features are quite similar. When HFR trained with MinMax is used to predict Kling, the accuracy improves by 18.7\%, and the AUC improves by 8.9\%, compared to CogVideoX-2B. However, when HFR trained with Kling is used to predict MinMax, the performance metrics show a slight decline relative to CogVideoX-2B.

Based on an analysis of the synthetic video sources from Kling and MinMax, we speculate that while both exhibit visual consistency and logical patterns close to real-world videos, Kling has a grainy texture, making its video quality noticeably inferior to that of MinMax. This explains the significant improvement in metrics when MinMax is used to predict Kling, while using Kling to predict MinMax shows minimal change. Future work could explore experiments involving mixed synthetic data sources for detection.

\section{Limitations}
\label{sec:lim}
The rapid evolution of video generation models poses a fundamental challenge to the sustainability of existing detection frameworks. As generative techniques advance continuously, current artifact detection mechanisms may become outdated quickly, necessitating perpetual updates and adaptive strategies to maintain robustness against novel forgery patterns.

\begin{figure*}
  \centering
\includegraphics[width=0.95\linewidth]{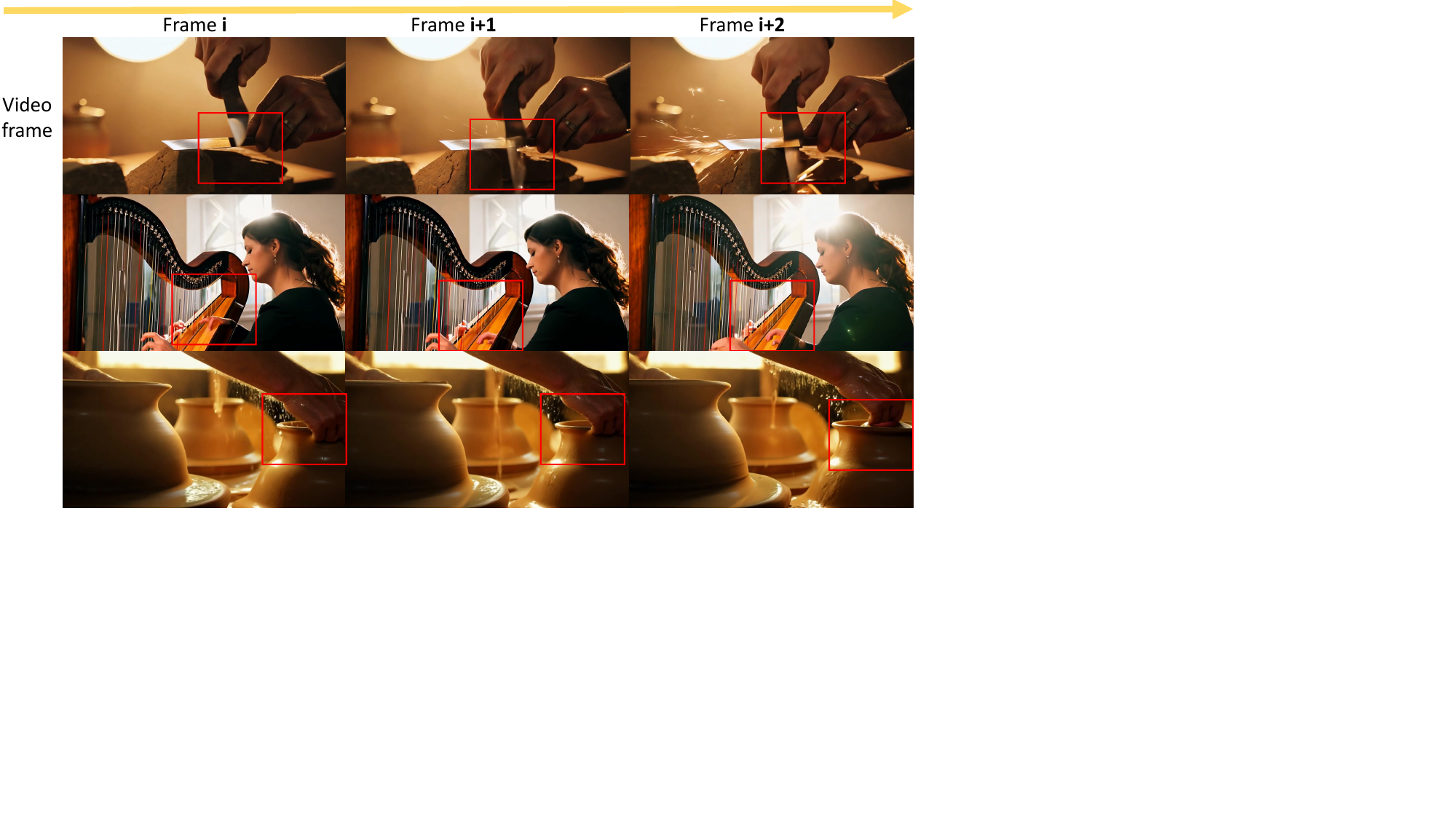}
  \caption{It can be seen that in the first row, two metal knives blur and pass through each other; in the second row, a woman's hand blurs as it reaches into the harp; in the third row, a person's hand passes through a clay pot under production without leaving any traces. Overall, this violates spatial logic and the normal rules of object interaction.}
  \label{fig:spatital}
\end{figure*}

\end{document}